\documentclass{midl} 

\usepackage{amsmath,amssymb}
\usepackage{caption}

\jmlrproceedings{}{}
\jmlrpages{}
\jmlryear{}

\jmlrworkshop{}
\jmlrvolume{}

\title[Stain Isolation-based Guidance for Stain Translation]{Stain Isolation-based Guidance \\for Improved Stain Translation}

\midlauthor{\Name{Nicolas Brieu} \Email{nicolas.brieu1@astrazeneca.com}\\
\Name{Felix J. Segerer} \Email{felix.segerer@astrazeneca.com}\\
\Name{Ansh Kapil} \Email{ansh.kapil@astrazeneca.com}\\
\Name{Philipp Wortmann} \Email{philipp.wortmann@astrazeneca.com}\\
\Name{Guenter Schmidt} \Email{guenter.schmidt@astrazeneca.com}\\
\addr{AstraZeneca Computational Pathology, Oncology R\&D, Munich, Germany}
}

\begin{document}

\maketitle

\vspace{-5mm}
\begin{abstract}
Unsupervised and unpaired domain translation using generative adversarial neural networks, and more precisely CycleGAN, is state of the art for the stain translation of histopathology images. It often, however, suffers from the presence of cycle-consistent but non structure-preserving errors. We propose an alternative approach to the set of methods which, relying on segmentation consistency, enable the preservation of pathology structures. Focusing on immunohistochemistry (IHC) and multiplexed immunofluorescence (mIF), we introduce a simple yet effective guidance scheme as a loss function that leverages the consistency of stain translation with stain isolation. Qualitative and quantitative experiments show the ability of the proposed approach to improve translation between the two domains. 
\end{abstract}

\begin{keywords}
Digital Pathology, GANs, Guided Domain Translation, Immunofluorescence
\end{keywords}

\vspace{-1mm}
\section{Introduction}
The computational analysis of mIF histopathological images, by enabling simultaneous and quantitative analysis of multiple markers, is becoming key for the development of novel therapeutic drugs. Because most deep-learning based segmentation and detection methods require large datasets of pixel-precise annotation to yield accurate and robust results, methods to transfer annotations or models from one domain to another are of increasing interest \cite{brieu2019}. However, the standard CycleGAN approach \cite{zhu2017} for the underlying unsupervised and unpaired domain translation have limitations in preservation of pathology structures, which led to the introduction of segmentation-based guidance \cite{mahapatra2020}. We propose in this study a novel and simple guidance scheme tailored to the translation from and to the IF domain, based on the conversion from the RGB to Haematoxylin-DAB (HD) colorspace \cite{ruifrok2001}. As shown qualitatively and quantitatively, our approach yields improved translation results.

\vspace{-3mm}
\section{Methods}
We incorporate two losses $\mathcal{L}^{IF}$ and $\mathcal{L}^{IHC}$ to the CycleGAN training (cf. Fig.~1). These respectively guide the translation from the IHC domain to the IF domain by generator $G_{AB}$ and the translation from the IF domain to the IHC domain by generator $G_{BA}$. 
With $x_{IHC}$ an IHC patch input to $\mathcal{G}_{AB}$ and $x_{IF}$ an IF patch input to $\mathcal{G}_{BA}$, we define the stain-isolation guidance loss $\mathcal{L}_g^{IF}$ as the $\mathcal{L}_1$ norm between the DAB channel of the generated IF image $\mathcal{G}_{AB}(x_{IHC})$ and the DAB channel of the pseudo-IF image $\mathcal{F}(x_{IHC})$. The later is derived from the conversion of $x_{IHC}$ to the Haematoxylin-DAB (HD) colorspace followed by its rescaling\footnote{https://scikit-image.org/docs/stable/auto\_examples/color\_exposure/plot\_ihc\_color\_separation.html} using a set of representative IHC images as reference. The inaccuracy of the estimated Haematoxylin channel leads us to consider only the DAB channel into the loss:
\vspace{-1mm}
\begin{equation}
\mathcal{L}_g^{IF} = \left\|\mathcal{G}_{AB}(x_{IHC}).{DAB} - \mathcal{F}(x_{IHC}).{DAB}\right\|_1
\end{equation}
\vspace{-1mm}
We similarly define the pseudo-IHC loss $\mathcal{L}_g^{IHC}$ as the $\mathcal{L}_1$ norm between the generated IHC image $\mathcal{G}_{BA}(x_{IF})$ and the image $\mathcal{F}^{-1}(x_{IF})$ obtained by color conversion from HD to RGB:
\vspace{-1mm}
\begin{equation}
\mathcal{L}_g^{IHC} = \left\|\mathcal{G}_{BA}(x_{IF}) - \mathcal{F}^{-1}(x_{IF})\right\|_1
\end{equation}
\vspace{-1mm}
The stain isolation and pseudo-IHC losses are added to the following CycleGAN losses: a cycle consistency loss $\mathcal{L}_{cycle}$, two adversarial losses on the output of the generators $\mathcal{G}_{AB}$ and $\mathcal{G}_{BA}$, an identity loss $\mathcal{L}_{id}$ as well as an embedding loss $\mathcal{L}_{emb}$. The overall loss reads as:
\vspace{-1mm}
\begin{equation}
\mathcal{L} = \mathcal{L}_{GAN}^{AB} + \mathcal{L}_{GAN}^{BA} + \lambda_1 \mathcal{L}_{cycle} + \lambda_2 \mathcal{L}_{id} + \lambda_3 \mathcal{L}_{emb} + \lambda_4 \mathcal{L}_g^{IHC} + \lambda_5 \mathcal{L}_g^{IF}
\label{eq:loss}
\end{equation}
\vspace{-1mm}
with the following fixed weighting parameters: $\lambda_1=10$, $\lambda_2=2$ and $\lambda_3=10$. In the remaining of this study, the two parameters $\lambda_4$ and $\lambda_5$ are either switched off or set to 10.

\begin{figure}[t!]
	\centering
	\includegraphics[width=0.95\linewidth]{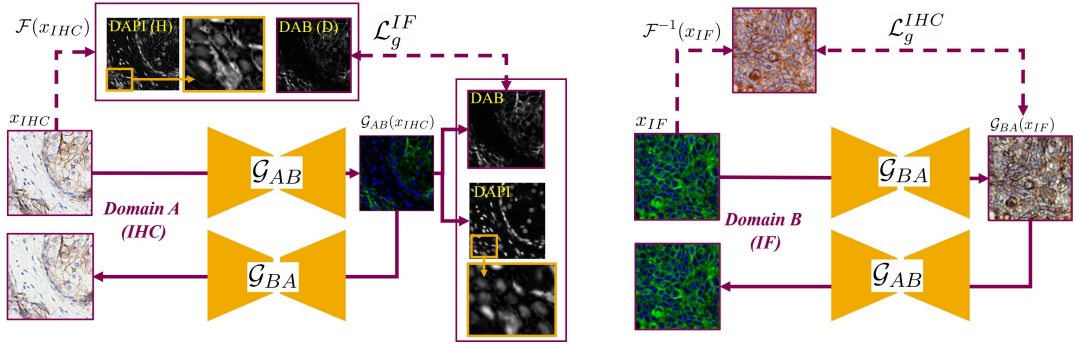}
	\caption{$\mathcal{L}_g^{IF}$ and $\mathcal{L}_g^{IHC}$ guide the training of $\mathcal{G}_{AB}$ and $\mathcal{G}_{BA}$. Note the improved DAPI channel of $\mathcal{G}_{AB}(x_{IHC})$ vs. $\mathcal{F}(x_{IHC})$ and the more realistic color of $\mathcal{G}_{BA}(x_{IF})$ vs. ${F}^{-1}(x_{IF})$.\vspace{-2mm}}
	\label{fig:methods}
\end{figure}

\vspace{-2mm}
\section{Results}
The unpaired IF and IHC datasets respectively contains 11K and 160K patches of $256\times256$px with $0.5\mu$m/px resolution. On the IF images, the PDL1 and DAPI channels are selected out of the seven initially available channels. The batch size  is 1, the learning rates 0.0001 for the generators and 0.0005 for the discriminators. Adam optimizer ($\beta_1=0.5$, $\beta_2=0.999$) is used to minimize the loss (cf. Eq.~\ref{eq:loss}) for 100K iterations. We qualitatively study the following configurations: (a) no guidance ($\lambda_4=0$, $\lambda_5=0$), (b) a virtual-IHC guidance ($\lambda_4=10$, $\lambda_5=0$), (c) a stain-isolation guidance ($\lambda_4=0$, $\lambda_5=10$) and (d) a combine guidance ($\lambda_4=10$, $\lambda_5=10$).
Fig.~2A shows the difficulty of baseline (a) to translate highly saturated regions: dark-brown pixels in IHC are wrongly translated into high DAPI signal in IF while some high DAPI signal in IF is wrongly translated into brown pixels in IHC. While these are partially corrected by the virtual-IHC loss (b), the stain-isolation loss (c) yields better translation. Combining both losses (d) does not improve over (c). This is quantitatively confirmed in Fig.~2B: proposed guidance schemes (b)-(d) prevent the generation of high DAPI signal in membrane regions, otherwise observed in (a).

\vspace{-2mm}
\section{Discussion}
We propose two novel loss functions based on color conversion in order to provide guidance at training time for the CycleGAN-based translation between the IF and IHC stain domains. Doing so, we are able to prevent cycle-consistent but non structure preserving translation errors. Future work include the inclusion of the proposed guidance losses to the second translation steps of each cycle as well as the downstream task of transferring segmentation and detection models from the IHC domain to the IF domain.

\begin{figure}[t!]
	\centering
	\includegraphics[width=0.99\linewidth]{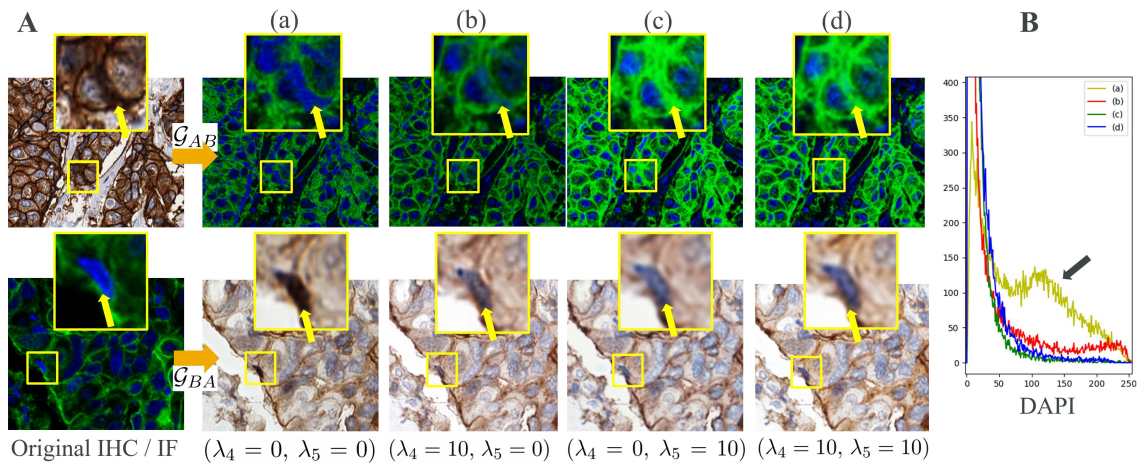}
	\caption{A - Examples of $\mathcal{G}_{AB}(x_{IHC})$ and $\mathcal{G}_{BA}(x_{IF})$ generated using: (a) the baseline CycleGAN, (b) the virtual-IHC guidance, (c) the stain-isolation guidance, and (d) combined guidance; B - Respective histograms of the DAPI values of $\mathcal{G}_{AB}(x_{IHC})$ measured on saturated ($S>0.90$) pixels of annotated membrane regions of 222 unseen IHC images $x_{IHC}$. \vspace{-2mm}}
\end{figure}

\vspace{-3mm}
\midlacknowledgments{\vspace{-1mm}We thank the mIF and IHC-HER2 teams of AstraZeneca Translation Medicine - Oncology R\&D, for their support in generating the image datasets used in this study.}

\vspace{-2mm}
\bibliography{midl-samplebibliography}

\begin{thebibliography}{4}
\providecommand{\natexlab}[1]{#1}
\providecommand{\url}[1]{\texttt{#1}}
\expandafter\ifx\csname urlstyle\endcsname\relax
  \providecommand{\doi}[1]{doi: #1}\else
  \providecommand{\doi}{doi: \begingroup \urlstyle{rm}\Url}\fi

\bibitem[Brieu(2019)]{brieu2019}
N.~Brieu.
\newblock Domain adaptation-based augmentation for weakly supervised nuclei
  detection.
\newblock In \emph{MICCAI workshop on Computational Pathology}, 2019.

\bibitem[Mahapatra(2020)]{mahapatra2020}
D.~Mahapatra.
\newblock Structure preserving stain normalization of histopathology images
  using self supervised semantic guidance.
\newblock In \emph{MICCAI}, 2020.

\bibitem[Ruifrok(2001)]{ruifrok2001}
A.~Ruifrok.
\newblock Quantification of histochemical staining by color deconvolution.
\newblock \emph{Analytical and quantitative cytology and histology}, 2001.

\bibitem[Zhu(2017)]{zhu2017}
J.~Zhu.
\newblock Unpaired image-to-image translation using cycle-consistent
  adversarial networks.
\newblock In \emph{ICCV}, 2017.

\end{thebibliography}

\let\clearpage\relax

\end{document}